\documentclass[11pt,a4paper]{article}
\usepackage[hyperref]{naaclhlt2018}
\usepackage{times}
\usepackage{latexsym}

\usepackage{booktabs} 
\usepackage{graphicx}

\usepackage{url}
\usepackage{epstopdf}
\usepackage{subfigure}
\usepackage{verbatim}
\usepackage{amssymb}
\usepackage{amsmath}
\usepackage{bm}
\usepackage[font={small}]{caption}
\usepackage{array}
\usepackage{multirow}

\newcommand{\nop}[1]{}
\newcommand{\rightarrowtext}[1]{$\xrightarrow{\text{#1}}$}
\newcommand{\leftarrowtext}[1]{$\xleftarrow{\text{#1}}$}

\newenvironment{remark}[1][Remark]{\begin{trivlist}
\item[\hskip \labelsep {\bfseries #1}]}{\end{trivlist}}

\newcolumntype{L}[1]{>{\raggedright\let\newline\\\arraybackslash\hspace{0pt}}m{#1}}
\newcolumntype{C}[1]{>{\centering\let\newline\\\arraybackslash\hspace{0pt}}m{#1}}
\newcolumntype{R}[1]{>{\raggedleft\let\newline\\\arraybackslash\hspace{0pt}}m{#1}}

\newenvironment{formalfont}{\fontfamily{pcr}\selectfont\small}{\par}
\DeclareTextFontCommand{\textformal}{\formalfont}

\aclfinalcopy 

\begin{document}
\title{Global Relation Embedding for Relation Extraction}

\author{Yu Su$^*$, Honglei Liu\thanks{$^*$ Equally contributed.}$\,\,$, Semih Yavuz, Izzeddin G{\" u}r \\ University of California, Santa Barbara \\ {\tt \{ysu,honglei,syavuz,izzeddingur\}@cs.ucsb.edu }
\AND
Huan Sun \\ The Ohio State University \\ {\tt sun.397@osu.edu }
\And
Xifeng Yan \\ University of California, Santa Barbara \\ {\tt xyan@cs.ucsb.edu }
}

\date{}

\maketitle

\begin{abstract}

We study the problem of textual relation embedding with distant supervision. To combat the wrong labeling problem of distant supervision, we propose to embed textual relations with \emph{global statistics} of relations, i.e., the co-occurrence statistics of textual and knowledge base relations collected from the entire corpus.  This approach turns out to be more robust to the training noise introduced by distant supervision. On a popular relation extraction dataset, we show that the learned textual relation embedding can be used to augment existing relation extraction models and significantly improve their performance. Most remarkably, for the top 1,000 relational facts discovered by the best existing model, the precision can be improved from 83.9\% to 89.3\%.

\end{abstract} 
\section{Introduction}
\label{sec:introduction}

\begin{figure*}[t]
    \includegraphics[width=.55\linewidth]{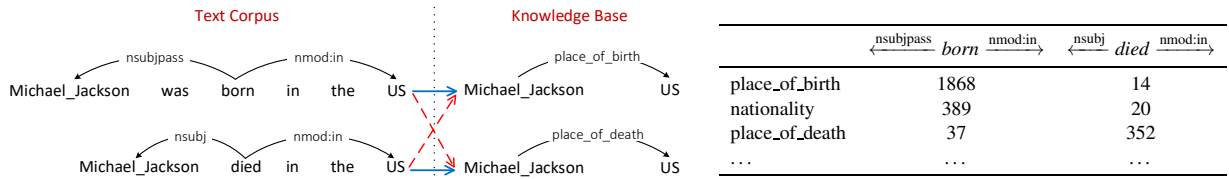}
    \quad
    \nop{
        \scalebox{0.7}{%
        \begin{tabular}[b]{c|l|l|l}
            \hline
            \textbf{Textual Relation}          &    \texttt{place\_of\_birth}    &   \texttt{nationality}    &   \texttt{place\_of\_death} \\
            \hline
            {\small \leftarrowtext{nsubjpass} \emph{born} \rightarrowtext{nmod:in}}   &  1868   &   389     &   37  \\
            {\small \leftarrowtext{nsubj} \emph{died} \rightarrowtext{nmod:in}}   & 14  &   20  &   352    \\
            \hline
        \end{tabular}
        }
    }
    \scalebox{0.68}{%
    \begin{tabular}[b]{lcc}
        \toprule
        &   \leftarrowtext{nsubjpass} \emph{born} \rightarrowtext{nmod:in}    &   \leftarrowtext{nsubj} \emph{died} \rightarrowtext{nmod:in}   \\
        \midrule
        place\_of\_birth   &   1868    &   14 \\
        nationality    &   389 &   20  \\
        place\_of\_death   &   37  &   352 \\
        \dots   &   \dots  &    \dots \\
        \bottomrule
    \end{tabular}
    }
    \caption{The wrong labeling problem of distant supervision, and how to combat it with global statistics. \emph{Left}: conventional distant supervision. Each of the textual relations will be labeled with both KB relations, while only one is correct (blue and solid), and the other is wrong (red and dashed). \emph{Right}: distant supervision with global statistics. The two textual relations can be clearly distinguished by their co-occurrence distribution of KB relations. Statistics are based on the annotated ClueWeb data released in \protect\cite{toutanova2015representing}.}
    \label{fig:wrong_label_problem}
\end{figure*}

Relation extraction requires deep understanding of the relation between entities.
Early studies mainly use hand-crafted features~\cite{kambhatla2004combining,guodong2005exploring}, and later kernel methods are introduced to automatically generate features~\cite{zelenko2003kernel,culotta2004dependency,bunescu2005shortest,zhang2006exploring}.
Recently neural network models have been introduced to embed words, relations, and sentences into continuous feature space, and have shown a remarkable success in relation extraction~\cite{socher2012semantic,zeng2014relation,xu2015classifying,zeng2015distant,lin2016neural}.
In this work, we study the problem of embedding \emph{textual relations}, defined as the shortest dependency path\footnote{We use fully lexicalized shortest dependency path with directional and typed dependency relations.} between two entities in the dependency graph of a sentence, to improve relation extraction.

Textual relations are one of the most discriminative textual signals that lay the foundation of many relation extraction models~\cite{bunescu2005shortest}.
A number of recent studies have explored textual relation embedding under the supervised setting~\cite{xu2015semantic,xu2015classifying,xu2016improved,liu2016relation}, but the reliance on supervised training data limits their scalability.
In contrast, we embed textual relations with \emph{distant supervision}~\cite{mintz2009distant}, which provides much larger-scale training data without the need of manual annotation.
However, the assertion of distant supervision, ``\emph{any} sentence containing a pair of entities that participate in a knowledge base (KB) relation is likely to express the relation,''  can be violated more often than not, resulting in many wrongly labeled training examples.
A representative example is shown in Figure~\ref{fig:wrong_label_problem}.
Embedding quality is thus compromised by the noise in training data.

Our main contribution is a novel way to combat the wrong labeling problem of distant supervision.
Traditional embedding methods~\cite{xu2015semantic,xu2015classifying,xu2016improved,liu2016relation} are based on \emph{local statistics}, i.e., individual textual-KB relation pairs like in Figure~\ref{fig:wrong_label_problem} (Left).
Our key hypothesis is that \emph{global statistics is more robust to noise than local statistics}.
For individual examples, the relation label from distant supervision may be wrong from time to time.
But when we zoom out to consider the entire corpus, and collect the global co-occurrence statistics of textual and KB relations, we will have a more comprehensive view of relation semantics:
The semantics of a textual relation can then be represented by its co-occurrence distribution of KB relations.
For example, the distribution in Figure~\ref{fig:wrong_label_problem} (Right) indicates that the textual relation {\small \textsc{Subject} \leftarrowtext{nsubjpass} \emph{born} \rightarrowtext{nmod:in} \textsc{Object}} mostly means {\small \texttt{place\_of\_birth}}, and is also a good indicator of {\small \texttt{nationality}}, but not {\small \texttt{place\_of\_death}}.
Although it is still wrongly labeled with {\small \texttt{place\_of\_death}} a number of times, the negative impact becomes negligible.
Similarly, we can confidently believe that {\small \textsc{Subject} \leftarrowtext{nsubj} \emph{died} \rightarrowtext{nmod:in} \textsc{Object}} means {\small \texttt{place\_of\_death}} in spite of the noise.
Textual relation embedding learned on such global statistics is thus more robust to the noise introduced by the wrong labeling problem.
\nop{Global statistics may also alleviate the impact of preprocessing errors (e.g., from dependency parser).}

We augment existing relation extractions using the learned textual relation embedding.
On a popular dataset introduced by Riedel et al.~\shortcite{riedel2010modeling}, we show that a number of recent relation extraction models, which are based on local statistics, can be greatly improved using our textual relation embedding.
Most remarkably, a new best performance is achieved when augmenting the previous best model with our relation embedding: The precision of the top 1,000 relational facts discovered by the model is improved from 83.9\% to 89.3\%, a 33.5\% decrease in error rate.
The results suggest that relation embedding with global statistics can capture complementary information to existing local statistics based models.

The rest of the paper is organized as follows.
In Section \ref{sec:related} we discuss related work.
For the modeling part, we first describe how to collect global co-occurrence statistics of relations in Section \ref{sec:relation_graph}, then introduce a neural network based embedding model in Section \ref{sec:embedding}, and finally discuss how to combine the learned textual relation embedding with existing relation extraction models in Section \ref{sec:relation_extraction}.
We empirically evaluate the proposed method in Section \ref{sec:evaluation}, and conclude in Section \ref{sec:conclusion}.

\section{Related Work}
\label{sec:related}

Relation extraction is an important task in information extraction.
Early relation extraction methods are mainly feature-based~\cite{kambhatla2004combining,guodong2005exploring}, where features in various levels, including POS tags, syntactic and dependency parses, are integrated in a max entropy model.
With the popularity of kernel methods, a large number of kernel-based relation extraction methods have been proposed~\cite{zelenko2003kernel,culotta2004dependency,bunescu2005shortest,zhang2006exploring}.
The most related work to ours is by Bunescu and Mooney~\cite{bunescu2005shortest}, where the importance of shortest dependency path for relation extraction is first validated.

More recently, relation extraction research has been revolving around neural network models, which can alleviate the problem of exact feature matching of previous methods and have shown a remarkable success (e.g.,~\cite{socher2012semantic,zeng2014relation}).
Among those, the most related are the ones embedding shortest dependency paths with neural networks~\cite{xu2015semantic,xu2015classifying,xu2016improved,liu2016relation}.
For example, Xu et al.~\shortcite{xu2015classifying} use a RNN with LSTM units to embed shortest dependency paths without typed dependency relations, while a convolutional neural network is used in~\cite{xu2015semantic}.
However, they are all based on the supervised setting with a limited scale.
In contrast, we embed textual relations with distant supervision~\cite{mintz2009distant}, which provides much larger-scale training data at a low cost.

Various efforts have been made to combat the long-criticized wrong labeling problem of distant supervision.
Riedel et al.~\shortcite{riedel2010modeling}, Hoffmann et al.~\shortcite{hoffmann2011knowledge}, and Surdeanu et al.~\shortcite{surdeanu2012multi} have attempted a multi-instance learning~\cite{dietterich1997solving} framework to soften the assumption of distant supervision, but their models are still feature-based.
Zeng et al.~\shortcite{zeng2015distant} combine multi-instance learning with neural networks, with the assumption that at least one of the contextual sentences of an entity pair is expressing the target relation, but this will lose useful information in the neglected sentences.
Instead, Lin et al.~\shortcite{lin2016neural} use all the contextual sentences, and introduce an attention mechanism to weight the contextual sentences.
Li et al.~\shortcite{ji2017distant} also use an attention mechanism to weight contextual sentences, and incorporate additional entity description information from knowledge bases.
Luo et al.~\shortcite{DBLP:conf/acl/LuoFWZHYZ17} manage to alleviate the negative impact of noise by modeling and learning noise transition patterns from data.
Liu et al.~\shortcite{liu2017heterogeneous} propose to infer the true label of a context sentence using a truth discovery approach \cite{li2016survey}.
Wu et al.~\shortcite{wu2017adversarial} incorporate adversarial training, i.e., injecting random perturbations in training, to improve the robustness of relation extraction.
Using PCNN+ATT \cite{lin2016neural} as base model, they show that adversarial training can improve its performance by a good margin.
However, the base model implementation used by them performed inferior to the one in the original paper and in ours, and therefore the results are not directly comparable.
No prior study has exploited global statistics to combat the wrong labeling problem of distant supervision.
Another unique aspect of this work is that we focus on compact textual relations, while previous studies along this line have focused on whole sentences.

In universal schema~\cite{riedel2013relation} for KB completion and relation extraction as well as its extensions~\cite{toutanova2015representing,verga2016multilingual}, a binary matrix is constructed from the entire corpus, with entity pairs as rows and textual/KB relations as columns.
A matrix entry is 1 if the relational fact is observed in training, and 0 otherwise.
Embeddings of entity pairs and relations, either directly or via neural networks, are then learned on the matrix entries, which are still individual relational facts, and the wrong labeling problem remains.
Global co-occurrence frequencies (see Figure~\ref{fig:wrong_label_problem} (Right)) are not taken into account, which is the focus of this study.
Another distinction is that our method directly models the association between textual and KB relations, while universal schema learns embedding for shared entity pairs and use that as a bridge between the two types of relations. 
It is an interesting venue for future research to comprehensively compare these two modeling approaches.

\section{Global Statistics of Relations}
\label{sec:relation_graph}

When using a corpus to train statistical models, there are two levels of statistics to exploit: \emph{local} and \emph{global}.
Take word embedding as an example.
The skip-gram model~\cite{mikolov2013distributed} is based on local statistics: During training, we sweep through the corpus and slightly tune the embedding model in each local window (e.g., 10 consecutive words).
In contrast, in global statistics based methods, exemplified by latent semantic analysis~\cite{deerwester1990indexing} and GloVe~\cite{pennington2014glove}, we process the entire corpus to collect global statistics like word-word co-occurrence counts, normalize the raw statistics, and train an embedding model directly on the normalized global statistics.

\begin{figure}[t]
\centering
\includegraphics[width=6cm]{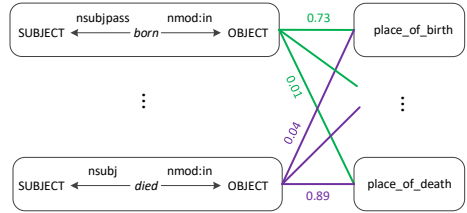}
\caption{Relation graph. The left node set is textual relations, and the right node set is KB relations. The raw co-occurrence counts are normalized such that the KB relations corresponding to the same textual relation form a valid probability distribution. Edges are colored by textual relation and weighted by normalized co-occurrence statistics.}
\label{fig:bi_graph}
\end{figure}

Most existing studies on relation extraction are based on local statistics of relations, i.e., models are trained on individual relation examples.
In this section, we describe how we collect global co-occurrence statistics of textual and KB relations, and how to normalize the raw statistics.
By the end of this section a bipartite \emph{relation graph} like Figure~\ref{fig:bi_graph} will be constructed, with one node set being textual relations $\mathcal{T}$, and the other being KB relations $\mathcal{R}$.
The edges are weighted by the normalized co-occurrence statistics of relations.

\subsection{Relation Graph Construction}

Given a corpus and a KB, we first do entity linking on each sentence, and do dependency parsing if at least two entities are identified\footnote{In the experiments entity linking is assumed given, and dependency parsing is done using Stanford Parser~\cite{chen2014fast} with universal dependencies.}.
For each entity pair $\left(e, e'\right)$ in the sentence, we extract the fully lexicalized shortest dependency path as a textual relation $t$, forming a \emph{relational fact} $\left(e, t, e'\right)$.
There are two outcomes from this step: a set of textual relations $\mathcal{T} = \left\{  t_i \right\}$, and the \emph{support} $S(t_i)$ for each $t_i$.
The support of a textual relation is a \emph{multiset} containing the entity pairs of the textual relation.
The \emph{multiplicity} of an entity pair, $m_{S(t_i)}( e, e' )$, is the number of occurrences of the corresponding relational fact $(e, t_i, e')$ in the corpus.
For example, if the support of $t_i$ is $S(t_i) = \left\{  \left(e_1, e'_1\right), \left(e_1, e'_1\right), \left(e_2, e'_2\right), \dots \right\}$, entity pair $\left( e_1, e'_1 \right)$ has a multiplicity of 2 because the relational fact $\left( e_1, t_i, e'_1 \right)$ occur in two sentences.
We also get a set of KB relations $\mathcal{R} = \left\{ r_j \right\}$, and the support $S(r_j)$ of a KB relation $r_j$ is the set of entity pairs having this relation in the KB, i.e., there is a relational fact $(e, r_j, e')$ in the KB.
The number of \emph{co-occurrences} of a textural relation $t_i$ and a KB relation $r_j$ is

\begin{equation}
n_{ij} = \sum_{\left(e, e'\right) \in S(r_j)} m_{S(t_i)}(e, e'),
\end{equation}

\noindent i.e., every occurrence of relational fact $(e, t_i, e')$ is counted as a co-occurrence of $t_i$ and $r_j$ if $\left(e, e'\right) \in S(r_j)$.
A bipartite relation graph can then be constructed, with $\mathcal{T}$ and $\mathcal{R}$ as the node sets, and the edge between $t_i$ and $r_j$ has weight $n_{ij}$ (no edge if $n_{ij} = 0$), which will be normalized later.

\subsection{Normalization}

The raw co-occurrence counts have a heavily skewed distribution that spans several orders of magnitude: A small portion of relation pairs co-occur highly frequently, while most relation pairs co-occur only a few times.
For example, a textual relation,
{\small \textsc{Subject} \leftarrowtext{nsubjpass} \emph{born} \rightarrowtext{nmod:in} \textsc{Object}},
may co-occur with the KB relation {\small \texttt{place\_of\_birth}} thousands of times (e.g., ``\emph{Michelle Obama was born in Chicago}''), while a synonymous but slightly more compositional textual relation,
{\small \textsc{Subject} \leftarrowtext{nsubjpass} \emph{born} \rightarrowtext{nmod:in} \emph{city} \rightarrowtext{nmod:of} \textsc{Object}},
may only co-occur with the same KB relation a few times in the entire corpus (e.g., ``\emph{Michelle Obama was born in the city of Chicago}'').
Learning directly on the raw co-occurrence counts, an embedding model may put a disproportionate amount of weight on the most frequent relations, and may not learn well on the majority of rarer relations.
Proper normalization is therefore necessary, which will encourage the embedding model to learn good embedding not only for the most frequent relations, but also for the rarer relations.

A number of normalization strategies have been proposed in the context of word embedding, including correlation- and entropy-based normalization~\cite{rohde2006improved}, positive pointwise mutual information (PPMI) \cite{bullinaria2007extracting}, and some square root type transformation~\cite{lebret2014word}.
A shared goal is to reduce the impact of the most frequent words, e.g., ``the'' and ``is,'' which tend to be less informative for the purpose of embedding.

We have experimented with a number of normalization strategies and found that the following strategy works best for textual relation embedding: For each textual relation, we normalize its co-occurrence counts to form a probability distribution over KB relations.
The new edge weights of the relation graph thus become $w_{ij} = \tilde{p}(r_j | t_i) = n_{ij} / \sum_{j'} n_{ij'}$.
Every textual relation is now associated with a set of edges whose weights sum up to 1.
We also experimented with PPMI and smoothed PPMI with $\alpha=0.75$ \cite{levy2015improving} that are commonly used in word embedding.
However, the learned textual relation embedding turned out to be not very helpful for relation extraction.
One possible reason is that PPMI (even the smoothed version) gives inappropriately large weights to rare relations \cite{levy2015improving}. 
There are many textual relations that correspond to none of the target KB relations but are falsely labeled with some KB relations a few times by distant supervision.
PPMI gives large weights to such falsely labeled cases because it thinks these events have a chance significantly higher than random. 

\section{Textual Relation Embedding}
\label{sec:embedding}

Next we discuss how to learn embedding of textual relations based on the constructed relation graph.
We call our approach \textbf{Glo}bal \textbf{R}elation \textbf{E}mbedding (GloRE) in light of global statistics of relations.

\subsection{Embedding via RNN}

Given the relation graph, a straightforward way of relation embedding is matrix factorization, similar to latent semantic analysis~\cite{deerwester1990indexing} for word embedding.
However, textual relations are different from words in that they are sequences composed of words and typed dependency relations.
Therefore, we use recurrent neural networks (RNNs) for embedding, which respect the compositionality of textual relations and can learn the shared sub-structures of different textual relations~\cite{toutanova2015representing}.
For the examples in Figure~\ref{fig:wrong_label_problem}, an RNN can learn, from both textual relations, that the  shared dependency relation ``{\small nmod:in}'' is indicative of location modifiers.
It is worth noting that other models like convolutional neural networks can also be used, but it is not the focus of this paper to compare all the alternative embedding models; rather, we aim to show the effectiveness of global statistics with a reasonable embedding model.

For a textual relation, we first decompose it into a sequence of tokens $\{x_1, ..., x_m\}$, which includes lexical words and directional dependency relations.
For example, the textual relation
{\small \textsc{Subject} \leftarrowtext{nsubjpass} \emph{born} \rightarrowtext{nmod:in} \textsc{Object}}
is decomposed to a sequence of three tokens \{{\small $-$nsubjpass, born, nmod:in}\}, where ``{\small $-$}'' represents a left arrow.
Note that we include directional dependency relations, because both the relation type and the direction are critical in determining the meaning of a textual relation.
For example, the dependency relation ``{\small nmod:in}'' often indicates a location modifier and is thus strongly associated with location-related KB relations like {\small \texttt{place\_of\_birth}}.
The direction also plays an important role.
Without knowing the direction of the dependency relations, it is impossible to distinguish {\small \texttt{child\_of}} and {\small \texttt{parent\_of}}.

\begin{figure}[t]
\centering
\includegraphics[width=7cm]{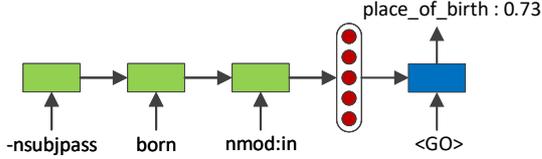}
\caption{Embedding model. \emph{Left}: A RNN with GRU for embedding. \emph{Middle}: embedding of textual relation. \emph{Right}: a separate GRU cell to map a textual relation embedding to a probability distribution over KB relations.}
\label{fig:rnn}
\end{figure}

An RNN with gated recurrent units (GRUs)~\cite{cho2014learning} is then applied to consecutively process the sequence as shown in Figure \ref{fig:rnn}.
We have also explored more advanced constructs like attention, but the results are similar, so we opt for a vanilla RNN in consideration of model simplicity.

Let $\phi$ denote the function that maps a token $x_l$ to a fixed-dimensional vector, the hidden state vectors of the RNN are calculated recursively:

\begin{equation}
\bm{h}_l = \textsc{GRU} \big ( \phi(x_l), \bm{h}_{l-1} \big ).
\end{equation}

\noindent \textsc{GRU} follows the definition in Cho et al.~\shortcite{cho2014learning}.

\subsection{Training Objective}
\label{sec:loss}

We use global statistics in the relation graph to train the embedding model.
Specifically, we model the semantics of a textual relation as its co-occurrence distribution of KB relations, and learn textual relation embedding to reconstruct the corresponding co-occurrence distributions.

We use a separate GRU cell followed by softmax to map a textual relation embedding to a distribution over KB relations; the full model thus resembles the sequence-to-sequence architecture~\cite{sutskever2014sequence}.
Given a textual relation $t_i$ and its embedding $\bm{h}_m$, the predicted conditional probability of a KB relation $r_j$ is thus:

\begin{equation}
p(r_j | t_i) = \text{softmax}( \textsc{GRU} ( \phi(\text{\textless GO\textgreater}), \bm{h}_{m} ) )_j,
\label{eq:prob}
\end{equation}

\noindent where $()_j$ denotes the $j$-th element of a vector, and \textless GO\textgreater is a special token indicating the start of decoding. The training objective is to minimize

\begin{equation}
\Theta = \frac{1}{|\mathcal{E}|} \sum_{i, j: \tilde{p}(r_j | t_i) > 0} \left(\log p(r_j|t_i) - \log \tilde{p}(r_j | t_i)  \right)^2,
\end{equation}

\noindent where $\mathcal{E}$ is the edge set of the relation graph.
It is modeled as a regression problem, similar to GloVe~\cite{pennington2014glove}.

\begin{remark}[Baseline.]
We also define a baseline approach where the unnormalized co-occurrence counts are directly used.
The objective is to maximize:

\begin{equation}
\Theta' = \frac{1}{\sum_{i, j} n_{ij}} \sum_{i, j: n_{ij} > 0} n_{ij} \log p(r_j|t_i).
\end{equation}

\noindent It also corresponds to local statistics based embedding, i.e., when the embedding model is trained on individual occurrences of relational facts with distant supervision.
Therefore, we call it \textbf{Lo}cal \textbf{R}elation \textbf{E}mbedding (LoRE).

\end{remark}

\section{Augmenting Relation Extraction}
\label{sec:relation_extraction}

Learned from global co-occurrence statistics of relations, our approach provides semantic matching information of textual and KB relations, which is often complementary to the information captured by existing relation extraction models.  In this section we discuss how to combine them together to achieve better relation extraction performance.

We follow the setting of distantly supervised relation extraction.
Given a text corpus and a KB with relation set $\mathcal{R}$, the goal is to find new relational facts from the text corpus that are not already contained in the KB.
More formally, for each entity pair $(e, e')$ and a set of \emph{contextual sentences} $C$ containing this entity pair, a relation extraction model assigns a score $E(z | C)$ to each candidate relational fact $z = (e, r, e'), r \in \mathcal{R}$.
On the other hand, our textual relation embedding model works on the sentence level.
It assign a score $G(z | s)$ to each contextual sentence $s$ in $C$ as for how well the textual relation $t$ between the entity pair in the sentence matches the KB relation $r$, i.e., $G(z | s) = p(r | t)$.
It poses a challenge to aggregate the sentence-level scores to get a set-level score $G(z | C)$, which can be used to combine with the original score $E(z | C)$ to get a better evaluation of the candidate relational fact.

One straightforward aggregation is max pooling, i.e., only using the largest score $\max_{s \in C} G(z|s)$, similar to the at-least-one strategy used by Zeng et al.~\shortcite{zeng2015distant}.
But it will lose the useful signals from those neglected sentences~\cite{lin2016neural}.
Because of the wrong labeling problem, mean pooling is problematic as well.
The wrongly labeled contextual sentences tend to make the aggregate scores more evenly distributed and therefore become less informative.
The number of contextual sentences positively supporting a relational fact is also an important signal, but is lost in mean pooling.

Instead, we use summation with a trainable $cap$:

\begin{equation}\label{eq:summation}
G(z|C) = \min{(cap, \sum_{s \in C}{G(z | s)} )},
\end{equation}

\noindent In other words, we additively aggregate the signals from all the contextual sentences, but only to a bounded degree.

We simply use a weighted sum to combine $E(z|C)$ and $G(z|C)$, where the trainable weights will also handle the possibly different scale of scores generated by different models:

\begin{equation}\label{eq:merge}
  \tilde{E}(z|C) = w_1 E(z | C) + w_2 G(z | C).
\end{equation}

\noindent The original score $E(z|C)$ is then replaced by the new score $\tilde{E}(z|C)$. To find the optimal values for $w_1$, $w_2$ and $cap$, we define a hinge loss:

\begin{equation}
\Theta_{\scriptscriptstyle Merge} = \frac{1}{K} \sum_{k=1}^{K} \max\big\{0, 1 + \tilde{E}(z_k^-) - \tilde{E}(z_k^+)\big\},
\label{eq:merge_loss}
\end{equation}

\noindent where $\{z_k^+\}_{k=1}^K$ are the true relational facts from the KB, and $\{z_k^-\}_{k=1}^K$ are false relational facts generated by replacing the KB relation in true relational facts with incorrect KB relations.

\section{Experiments}
\label{sec:evaluation}

In this experimental study, we show that GloRE can greatly improve the performance of several recent relation extraction models, including the previous best model on a standard dataset.

\subsection{Experimental Setup}

\begin{remark}[Dataset.]

Following the literature~\cite{hoffmann2011knowledge,surdeanu2012multi,zeng2015distant,lin2016neural},
we use the relation extraction dataset introduced in~\cite{riedel2010modeling}, which was generated by aligning New York Times (NYT) articles with Freebase~\cite{bollacker2008freebase}.
Articles from year 2005-2006 are used as training, and articles from 2007 are used as testing.
Some statistics are listed in Table \ref{tab:stat}.
There are 53 target KB relations, including a special relation NA indicating that there is no target relation between entities.

\begin{table}[t]
\centering
\small
\scalebox{0.77}{
\begin{tabular}{cccc}
\toprule
Data & \# of sentences  & \# of entity pairs & \# of relational facts from KB\\
\midrule
Train & 570,088 & 291,699 & 19,429  \\
Test & 172,448  & 96,678 & 1,950 \\
\bottomrule
\end{tabular}
}
\caption{Statistics of the NYT dataset.}\label{tab:stat}
\end{table}

We follow the approach described in Section~\ref{sec:relation_graph} to construct the relation graph from the NYT training data.
The constructed relation graph contains 321,447 edges with non-zero weight.
We further obtain a training set and a validation set from the edges of the relation graph.
We have observed that using a validation set totally disjoint from the training set leads to unstable validation loss, so we randomly sample 300K edges as the training set, and another 60K as the validation set.
The two sets can have some overlap.
For the merging model (Eq.~\ref{eq:merge_loss}), 10\% of the edges are reserved as the validation set.

\end{remark}

\begin{figure*}[t]
\centering
\subfigure{
\label{fig:pr_cnn_att}
\includegraphics[width=5cm]{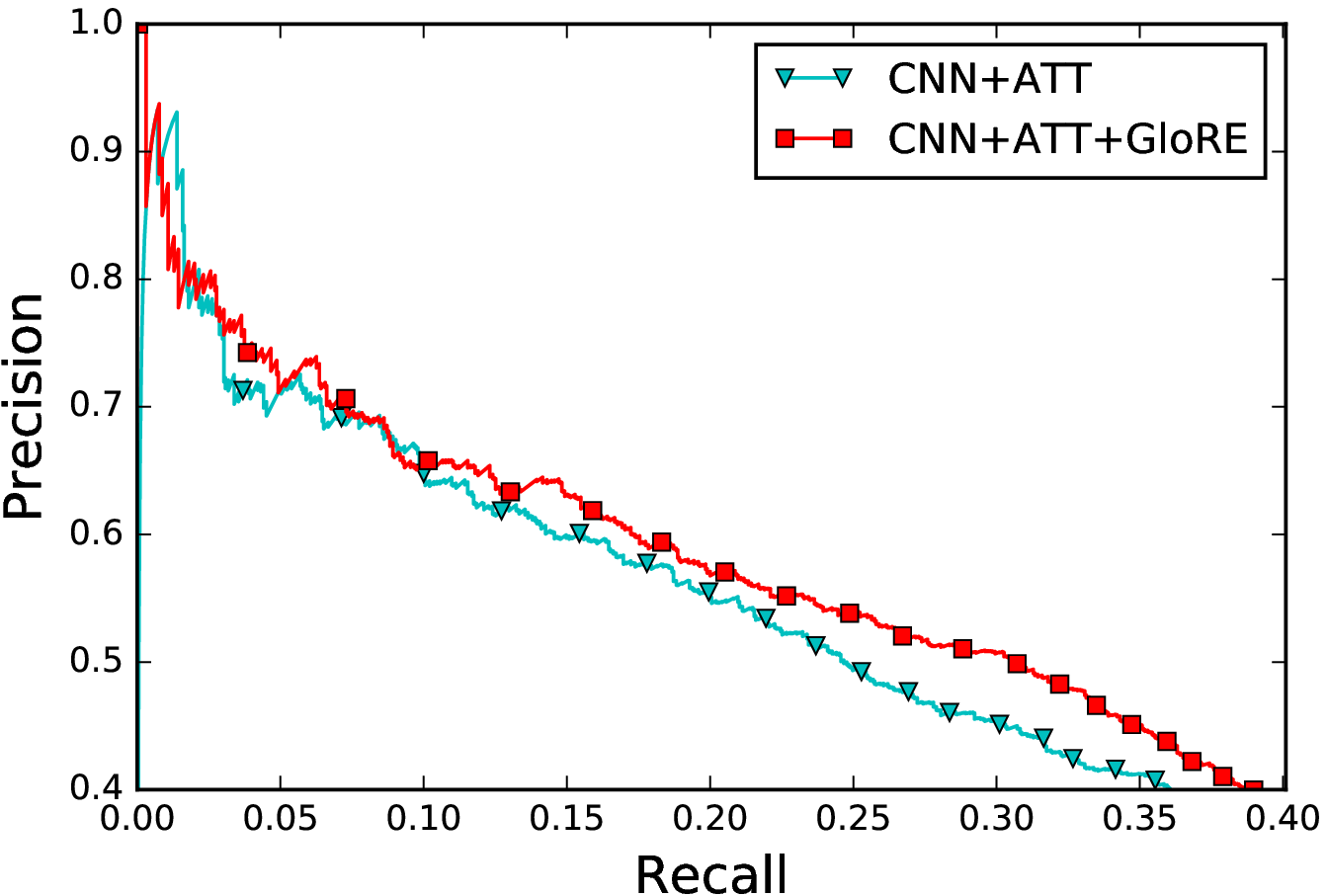}
}
\subfigure{
\label{fig:pr_cnn_one}
\includegraphics[width=5cm]{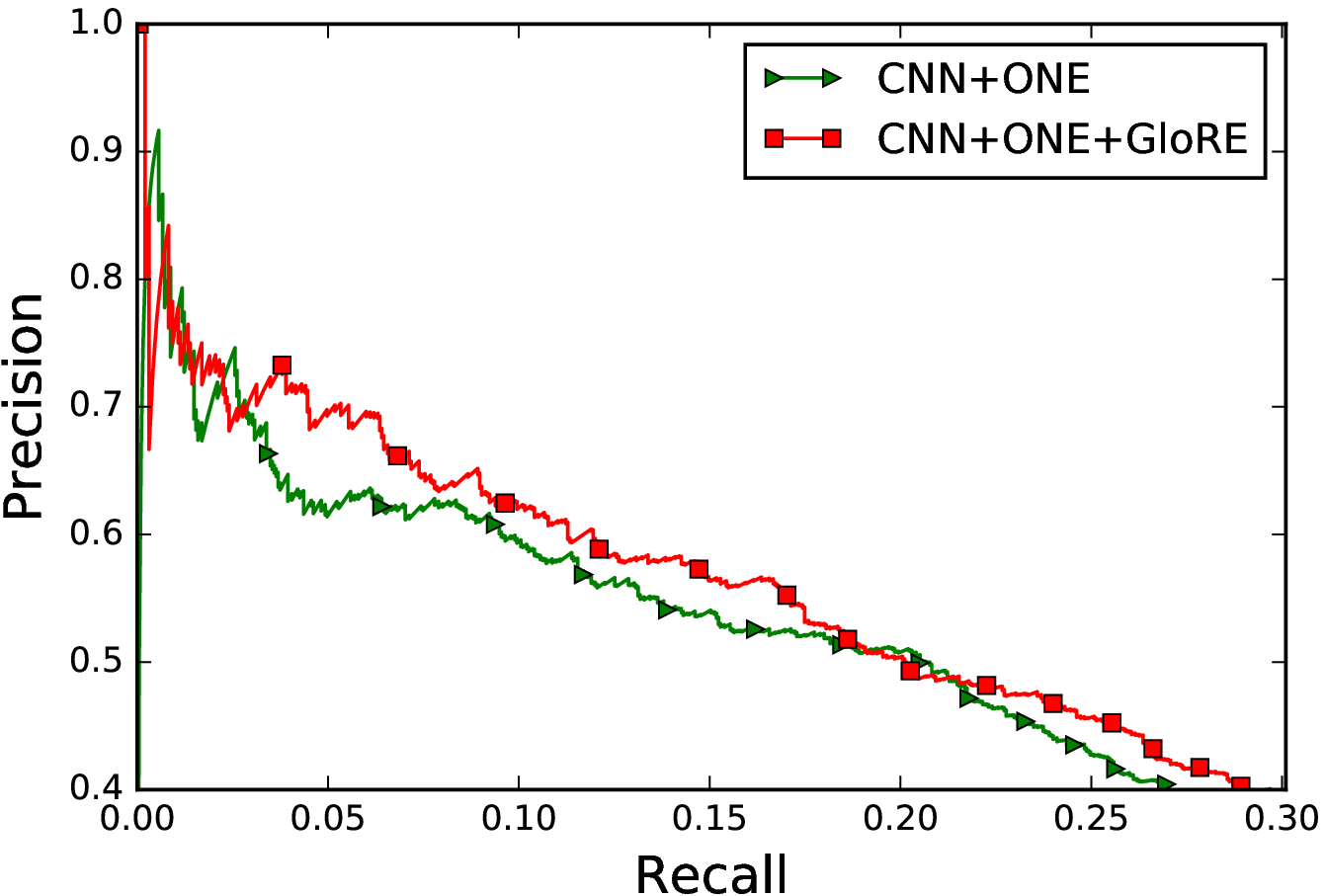}
}
\subfigure{
\label{fig:pr_pcnn_one}
\includegraphics[width=5cm]{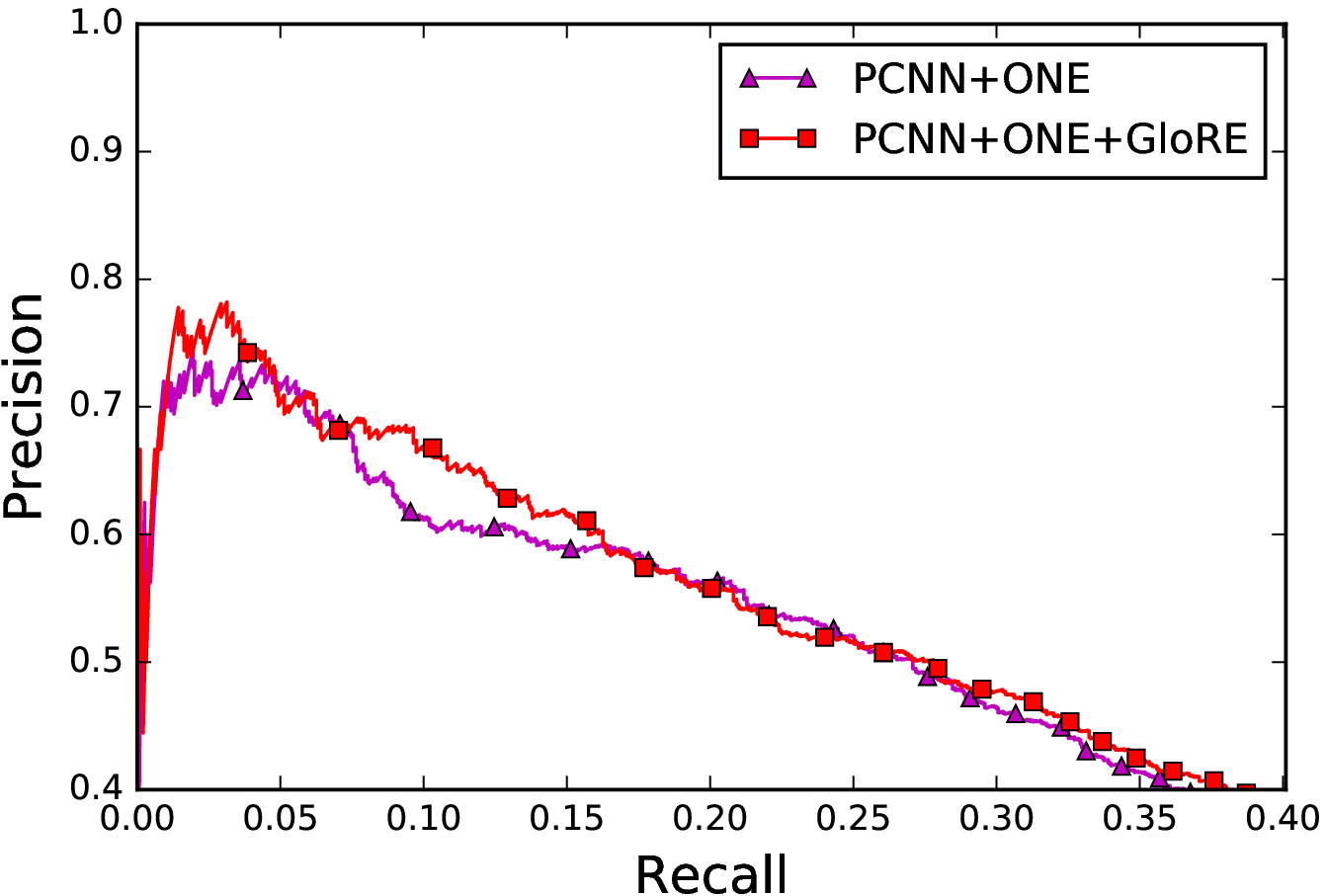}
}
\caption{Held-out evaluation: other base relation extraction models and the improved versions when augmented with GloRE.}
\label{fig:pr_sep}
\end{figure*}

\begin{figure}[t]
\centering
\includegraphics[width=.8\linewidth]{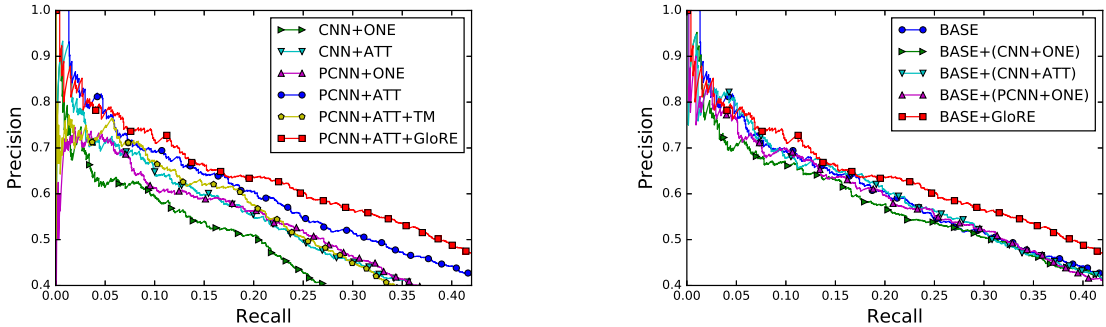}
\caption{Held-out evaluation: the previous best-performing model can be further improved when augmented with GloRE. PCNN+ATT+TM is a recent model~\cite{DBLP:conf/acl/LuoFWZHYZ17} whose performance is slightly inferior to PCNN+ATT. Because the source code is not available, we did not experiment to augment this model with GloRE. Another recent method \cite{wu2017adversarial} incorporates adversarial training to improve PCNN+ATT, but the results are not directly comparable (see Section \ref{sec:related} for more discussion). Finally, Ji et al.~\shortcite{ji2017distant} propose a model similar to PCNN+ATT, but the performance is inferior to PCNN+ATT and is not shown here for clarity.}
\label{fig:pr_all}
\end{figure}

\begin{figure}[t]
\centering
\includegraphics[width=.8\linewidth]{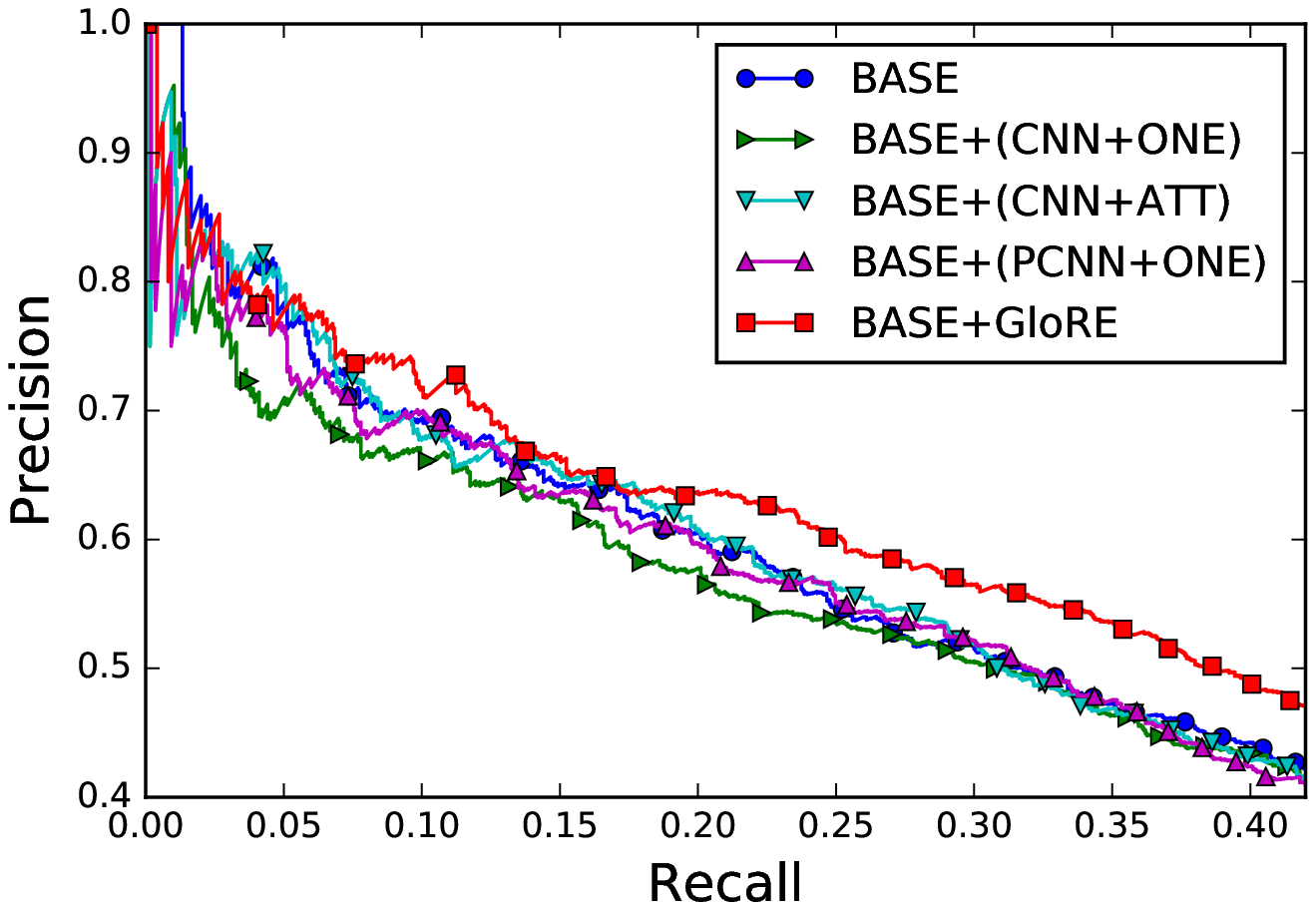}
\caption{Held-out evaluation: GloRE brings the largest improvement to BASE (PCNN+ATT), which further shows that GloRE captures useful information for relation extraction that is complementary to existing models.}
\label{fig:ensemble}
\end{figure}

\begin{remark}[Relation extraction models.]

We evaluate with four recent relation extraction models whose source code is publicly available\footnote{\url{https://github.com/thunlp/NRE}}.
We use the optimized parameters provided by the authors.

\begin{itemize}
  \item \textbf{CNN+ONE} and \textbf{PCNN+ONE}~\cite{zeng2015distant}: A convolutional neural network (CNN) is used to embed contextual sentences for relation classification. Multi-instance learning with at-least-one (ONE) assumption is used to combat the wrong labeling problem. In PCNN, piecewise max pooling is used to handle the three pieces of a contextual sentence (split by the two entities) separately.
  \item \textbf{CNN+ATT} and \textbf{PCNN+ATT}~\cite{lin2016neural}: Different from the at-least-one assumption which loses information in the neglected sentences, these models learn soft attention weights (ATT) over contextual sentences and thus can use the information of all the contextual sentences. \emph{PCNN+ATT is the best-performing model on the NYT dataset}.
\end{itemize}

\end{remark}

\begin{remark}[Evaluation settings and metrics.]

Similar to previous work~\cite{riedel2010modeling,zeng2015distant}, we use two settings for evaluation:
(1) Held-out evaluation, where a subset of relational facts in KB is held out from training (Table~\ref{tab:stat}), and is later used to compare against newly discovered relational facts.
This setting avoids human labor but can introduce some false negatives because of the incompleteness of the KB.
(2) Manual evaluation, where the discovered relational facts are manually judged by human experts.
For held-out evaluation, we report the precision-recall curve.
For manual evaluation, we report $Precision@N$, i.e., the precision of the top $N$ discovered relational facts.

\end{remark}

\begin{remark}[Implementation.]

Hyper-parameters of our model are selected based on the validation set.
For the embedding model, the mini-batch size is set to 128, and the state size of the GRU cells is 300.
For the merging model, the mini-batch size is set to 1024.
We use Adam with parameters recommended by the authors for optimization.
Word embeddings are initialized with the 300-dimensional word2vec vectors pre-trained on the Google News corpus\footnote{\url{https://code.google.com/archive/p/word2vec/}}.
Early stopping based on the validation set is employed.
Our model is implemented using Tensorflow~\cite{abadi2016tensorflow}, and the source code is available at \url{https://github.com/ppuliu/GloRE}.

\end{remark}

\subsection{Held-out Evaluation}

\begin{remark}[Existing Models + GloRE.]

We first show that our approach, GloRE, can improve the performance of the previous best-performing model, PCNN+ATT, leading to a new state of the art on the NYT dataset.
As shown in Figure \ref{fig:pr_all}, when PCNN+ATT is augmented with GloRE, a consistent improvement along the precision-recall curve is observed.
It is worth noting that although PCNN+ATT+GloRE seems to be inferior to PCNN+ATT when recall $< 0.05$, as we will show via manual evaluation, it is actually due to false negatives.

We also show in Figure \ref{fig:pr_sep} that the improvement brought by GloRE is general and not specific to PCNN+ATT; the other models also get a consistent improvement when augmented with GloRE.

To investigate whether the improvement brought by GloRE is simply from ensemble, we also augment PCNN+ATT with the other three base models in the same way as described in Section \ref{sec:relation_extraction}.
The results in Figure~\ref{fig:ensemble} show that pairwise ensemble of existing relation extraction models does not yield much improvement, and GloRE brings much larger improvement than the other models.

In summary, the held-out evaluation results suggest that GloRE captures useful information for relation extraction that is not captured by these local statistics based models.

\end{remark}

\begin{figure}[t]
\centering
\includegraphics[width=.8\linewidth]{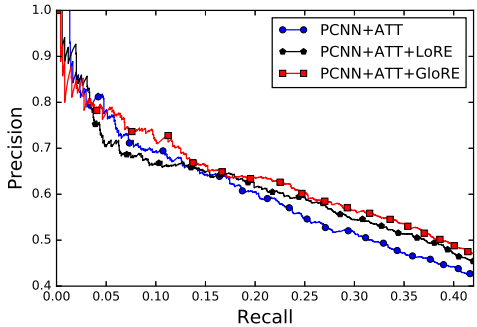}
\caption{Held-out evaluation: LoRE vs. GloRE.}
\label{fig:pr_local}
\end{figure}

\begin{remark}[LoRE v.s. GloRE.]

We compare GloRE with the baseline approach LoRE (Section~\ref{sec:embedding}) to show the advantage of normalization on global statistics.
We use PCNN+ATT as the base relation extraction model.
As shown in Figure~\ref{fig:pr_local}, GloRE consistently outperforms LoRE.
It is worth noting that LoRE can still improve the base relation extraction model when recall $> 0.15$, further confirming the usefulness of directly embedding textual relations in addition to sentences.

\end{remark}

\subsection{Manual Evaluation}

Due to the incompleteness of the knowledge base, held-out evaluation introduces some false negatives.
The precision from held-out evaluation is therefore a lower bound of the true precision.
To get a more accurate evaluation of model performance, we have human experts to manually check the false relational facts judged by held-out evaluation in the top 1,000 predictions of three models, PCNN+ATT, PCNN+ATT+LoRE and PCNN+ATT+GloRE, and report the corrected results in Table \ref{tab:maual}.
Each prediction is examined by two human experts who reach agreement with discussion.
To ensure fair comparison, the experts are not aware of the provenance of the predictions.
Under manual evaluation, PCNN+ATT+GloRE achieves the best performance in the full range of $N$.
In particular, for the top 1,000 predictions, GloRE improves the precision of the previous best model PCNN+ATT from 83.9\% to 89.3\%.
The manual evaluation results reinforce the previous observations from held-out evaluation.

\begin{table}[t]
\centering
\small
\scalebox{0.85}{
\begin{tabular}{lllllll}
\toprule
Precision@$N$ & 100  & 300 & 500 & 700 & 900 & 1000 \\
\midrule
PCNN+ATT  & \textbf{97.0} & 93.7 & 92.8 & 89.1 & 85.2 & 83.9 \\
PCNN+ATT+LoRE & \textbf{97.0}  & 95.0 & 94.2 & 91.6 & 89.6 & 87.0  \\
PCNN+ATT+GloRE & \textbf{97.0} & \textbf{97.3} & \textbf{94.6} & \textbf{93.3} & \textbf{90.1} & \textbf{89.3} \\
\bottomrule
\end{tabular}
}
\caption{Manual evaluation: false negatives from held-out evaluation are manually corrected by human experts.}\label{tab:maual}
\end{table}

\begin{table*}[t]
\centering
\footnotesize
\scalebox{0.9}{%
\begin{tabular}{lllll}
\toprule
Contextual Sentence & Textual Relation & PCNN+ATT Predictions & LoRE Predictions &  GloRE Predictions \\

\midrule

\multirow{3}{4.5cm}{[\textbf{Alfred Blumstein}]$_{head}$, a criminologist at [\textbf{Carnegie Mellon University}]$_{tail}$, called \dots} &
\multirow{3}{2.5cm}{\leftarrowtext{appos} criminologist  \rightarrowtext{nmod:at}} &
NA (0.63) &
\textbf{employee\_of} (1.00) &
\textbf{employee\_of} (0.96)
\\

& &
\textbf{employee\_of} (0.36) &
NA (0.00) &
NA (0.02)
\\

& &
founder\_of (0.00) &
founder\_of (0.00) &
founder\_of (0.02)
\\

\midrule

\multirow{3}{4.5cm}{[\textbf{Langston Hughes}]$_{head}$, the American poet, playwright and novelist, came to [\textbf{Spain}]$_{tail}$ to \dots} &
\multirow{3}{2.5cm}{\leftarrowtext{-nsubj} came  \rightarrowtext{   to   }} &
\textbf{NA} (0.58) &
place\_of\_death (0.35) &
\textbf{NA} (0.73)
\\

& &
nationality (0.38) &
\textbf{NA} (0.33) &
contain\_location (0.07)
\\

& &
place\_lived (0.01) &
nationality (0.21) &
employee\_of (0.06)
\\

\bottomrule
\end{tabular}
} %
\caption{Case studies. We select entity pairs that have only one contextual sentence, and the head and tail entities are marked. The top 3 predictions from each model with the associated probabilities are listed, with the correct relation bold-faced.}\label{tab:case_study}
\end{table*}

\subsection{Case Study}

Table~\ref{tab:case_study} shows two examples.
For better illustration, we choose entity pairs that have only one contextual sentence.

For the first example, PCNN+ATT predicts that most likely there is no KB relation between the entity pair, while both LoRE and GloRE identify the correct relation with high confidence.
The textual relation clearly indicates that the head entity is ({\small appos}) a criminologist at ({\small nmod:at}) the tail entity.

For the second example, there is no KB relation between the entity pair, and PCNN+ATT is indeed able to rank NA at the top.
However, it is still quite confused by \texttt{nationality}, probably because it has learned that sentences about a person and a country with many words about profession (``poet,'' ``playwright,'' and ``novelist'') likely express the person's nationality.
As a result, its prediction on NA is not very confident.
On the other hand, GloRE learns that if a person ``came to'' a place, likely it is not his/her birthplace.
In the training data, due to the wrong labeling problem of distant supervision, the textual relation is wrongly labeled with \texttt{place\_of\_death} and \texttt{nationality} a couple of times, and both PCNN+ATT and LoRE suffer from the training noise.
Taking advantage of global statistics, GloRE is more robust to such noise introduced by the wrong labeling problem.

\section{Conclusion}
\label{sec:conclusion}

Our results show that textual relation embedding trained on global co-occurrence statistics captures useful relational information that is often complementary to existing methods.
As a result, it can greatly improve existing relation extraction models.
Large-scale training data of embedding can be easily solicited from distant supervision, and the global statistics of relations provide a natural way to combat the wrong labeling problem of distant supervision.

The idea of relation embedding based on global statistics can be further expanded along several directions.
In this work we have focused on embedding textual relations, but it is in principle beneficial to jointly embed knowledge base relations and optionally entities.
Recently a joint embedding approach has been attempted in the context of knowledge base completion~\cite{toutanova2015representing}, but it is still based on local statistics, i.e., individual relational facts.
Joint embedding with global statistics remains an open problem.
Compared with the size of the training corpora for word embedding (up to hundred of billions of tokens), the NYT dataset is quite small in scale.
Another interesting venue for future research is to construct much larger-scale distant supervision datasets to train general-purpose textual relation embedding that can help a wide range of downstream relational tasks such as question answering and textual entailment.

\section*{Acknowledgements}

The authors would like to thank the anonymous reviewers for their thoughtful comments. This research
was sponsored in part by the Army Research Laboratory under cooperative agreements
W911NF09-2-0053 and NSF IIS 1528175. The views and conclusions contained herein are those
of the authors and should not be interpreted as representing the official policies, either expressed or
implied, of the Army Research Laboratory or the U.S. Government. The U.S. Government is authorized to reproduce and distribute reprints for Government
purposes notwithstanding any copyright notice herein.

\bibliography{glore}
\bibliographystyle{acl_natbib}

\end{document}